\pdfoutput=1
\pdfoutput=1
\pdfoutput=1
\pdfoutput=1
\pdfoutput=1

\documentclass[runningheads]{llncs}
\usepackage{graphicx}
\usepackage{mathrsfs}
\usepackage{amsmath,amssymb,amsfonts}
\usepackage{algorithm} 
\usepackage{algorithmic}
\usepackage{caption}
\usepackage{multirow}
\usepackage{xcolor}
\usepackage{enumitem}
\begin{document}
\title{A Tailored NSGA-III Instantiation for Flexible Job Shop Scheduling\thanks{This work is part of the research programme Smart Industry SI2016 with project name CIMPLO and project number 15465, which is (partly) financed by the Netherlands Organisation for Scientific Research (NWO).}}
%
\author{Yali Wang\and
Bas van Stein \and
Michael T.M. Emmerich \and
Thomas B{\"a}ck}

%
\authorrunning{Yali Wang et al.}
%
\institute{Leiden Institute of Advanced Computer Science, Leiden University, Niels Bohrweg 1, 2333CA Leiden, The Netherlands\\
\email{y.wang@liacs.leidenuniv.nl}\\
}
\maketitle              
\begin{abstract}
A customized multi-objective evolutionary algorithm (MOEA) is proposed for the multi-objective flexible job shop scheduling problem (FJSP). It uses smart initialization approaches to enrich the first generated population, and proposes various crossover operators to create a better diversity of offspring. Especially, the \emph{MIP-EGO configurator}, which can tune algorithm parameters, is adopted to automatically tune operator probabilities. Furthermore, different local search strategies are employed to explore the neighborhood for better solutions. In general, the algorithm enhancement strategy can be integrated with any standard EMO algorithm. In this paper, it has been combined with NSGA-III to solve benchmark multi-objective FJSPs, whereas an off-the-shelf implementation of NSGA-III is not capable of solving the FJSP. The experimental results show excellent performance with less computing budget.

\end{abstract}
\section{Introduction}

The Job shop scheduling problem (JSP) is an important branch of production planning problems. The classical JSP consists of a set of independent jobs to be processed on multiple machines and each job contains a number of operations with a predetermined order. It is assumed that each operation must be processed on a specific machine with a specified processing time. The JSP is to determine a schedule of jobs, meaning to sequence operations on the machines. The flexible job shop scheduling problem (FJSP) is an important extension of the classical JSP due to the wide employment of multi-purpose machines in the real-world job shop. The FJSP extends the JSP by assuming that each operation is allowed to be processed on a machine out of a set of alternatives, rather than one specified machine. Therefore, the FJSP is not only to find the best sequence of operations on a machine, but also to assign each operation to a machine out of a set of qualified machines. The JSP is well known to be strongly NP-hard \cite{garey1976complexity}. The FJSP is an even more complex version of the JSP, so the FJSP is clearly also strongly NP-hard.

A typical objective of the FJSP is the \emph{makespan}, which is defined as the maximum time for completion of all jobs, in other words, the total length of the schedule. However, to achieve a practical schedule for the FJSP, various conflicting objectives should be considered. In this paper, evolutionary algorithms (EA) have been applied to a multi-objective flexible job shop scheduling problem (MOFJSP) with three objectives, namely: The makespan, total workload and critical workload. We propose and adopt multiple initialization approaches to enrich the first generated population based on our definition of the chromosome representation; at the same time, diverse genetic operators are applied to guide the search towards offspring with a wide diversity; especially, we use an algorithm configurator to tune the parameter configuration; furthermore, two levels of local search are employed leading to better solutions. Our proposed FJSP multi-objective evolutionary algorithm (FJSP-MOEA) can be combined with almost all MOEAs to solve MOFJSP, the experimental results show that FJSP-MOEA can achieve the state-of-the-art results with less computational effort when we merge it with NSGA-III \cite{deb2013evolutionary}. 

The paper is organized as follows. The next section formulates the MOFJSP, which is the problem we are about to solve. Section~\ref{sec:relatedwork} gives necessary background knowledge. Section~\ref{sec:proposedalgorithm} introduces the proposed algorithm and Section~\ref{sec:experiments} reports the experimental results. Finally, Section~\ref{sec:conclusions} concludes the work and suggests future work directions.

\section{PROBLEM FORMULATION}
\label{sec:problem}

The MOFJSP addressed in this paper is described as follows:

\begin{enumerate}
    \item There are $n$ jobs $J=\{{J_{1},J_{2}},\cdots,J_{n}\}$ and $m$ machines $M=\{{M_{1},M_{2}},\cdots,M_{m}\}$.
    
    \item Each job $J_i$ comprises $l_i$ operations for $i=1,\cdots,n$, the $j$th operation of job $J_i$ is represented by $O_{ij}$, and the operation sequence of job $J_i$ is from $O_{i1}$ to $O_{il_i}$.

    \item For each operation $O_{ij}$, there is a set of machines capable of performing it, which is represented by $M_{ij}$ and it is a subset of $M$.
    
    \item The processing time of the operation $O_{ij}$ on machine $M_k$ is predefined and denoted by $t_{ijk}$.
\end{enumerate}

At the same time, the following assumptions are made:

\begin{enumerate}
    \item All machines are available at time $0$ and assumed to be continuously available.
    
    \item All jobs are released at time $0$ and independent from each other.
    
    \item Setting up times of machines and transportation times between operations are negligible.
    
    \item Environmental changes (such as machine breakdowns) are neglected.
     
    \item A machine can only work on one operation at a time.
    
    \item There are no precedence constraints among the operations of different jobs, and the order of operations for each job cannot be modified.
    
    \item An operation, once started, must run to completion.
    
    \item No operation for a job can be started until the previous operation for that job is completed.
\end{enumerate}

The makespan, total workload and critical workload, which are commonly considered in the literature on FJSP (e.g., \cite{chiang2013simple}, \cite{yuan2013multiobjective}), are minimized and used as the three objectives in our algorithm. Minimizing the makespan can facilitate the rapid response to the market demand. The total workload represents the total working time of all machines and the critical workload is the maximum workload among all machines. Minimizing the total workload can reduce the use of machines; minimizing the critical workload can balance the workload between machines. Let $C_i$ denote the completion time of job $J_i$, $W_k$ the sum of processing time of all operations that are processed on machine $M_k$. The three objectives can be defined as follows: 
\begin{flalign}
&\text{makespan} (C_{max}) :  f_1 = \max\{ C_i | i=1,2,\cdots, n\} \\
&\text{total workload} (W_t): f_2 = \sum_{k=1}^{m}W_k \\
&\text{critical workload} (W_{max}): f_3 = \max\{ W_k | k=1,2,\cdots, m\} 
\end{flalign}

An example of MOFJSP is shown in Table~\ref{table_1} as an illustration, where rows correspond to operations and columns correspond to machines. In this example, there are three machines: $M_1$, $M_2$ and $M_3$. Each entry of the table denotes the processing time of that operation on the corresponding machine, and the tag $``-"$ means that a machine cannot execute the corresponding operation.

\begin{table}[htbp]
\caption{Processing time of a FJSP instance}
\label{table_1}
\begin{center}
\begin{tabular}{c|c|c|c|c}
\hline
Job & Operation & $M_1$ & $M_2$ & $M_3$\\
\hline
\multirow{3}{*}{$J_1$} & $O_{11}$ & 3 & - & 2\\
 & $O_{12}$ & 5 & 7 & 6\\
 & $O_{13}$ & - & - & 2\\
\hline
\multirow{2}{*}{$J_2$} & $O_{21}$ & 2 & 4 & 3\\
 & $O_{22}$ & 2 & - & 1\\
\hline
\multirow{2}{*}{$J_3$} & $O_{31}$ & 4 & 2 & 2\\
 & $O_{32}$ & 3 & 5 & -\\
\hline
\end{tabular}
\end{center}
\end{table}

\section{RELATED WORK}
\label{sec:relatedwork}

\subsection{Algorithms for MOFJSP}
The FJSP has been investigated extensively in the last three decades. According to \cite{chaudhry2016research}, EA is the most popular non-hybrid technique to solve the FJSP. Among all EAs for FJSP, some are developed for the more challenging FJSP: the MOFJSP which we formulated in section 2. \cite{wang2010multi}, \cite{chiang2013simple} and \cite{yuan2013multiobjective} are very successful MOFJSP algorithms and have obtained high-quality solutions. \cite{wang2010multi} proposed a multi-objective genetic algorithm (MOGA) based on the immune and entropy principle. In this MOGA, the fitness was determined by the Pareto dominance relation and the diversity was kept by the immune and entropy principle. In \cite{chiang2013simple}, a simple EA (SEA) was proposed, which used domain heuristics to generate the initial population and balanced the exploration and exploitation by refining duplicate individuals with mutation operators. A memetic algorithm (MA) was proposed in \cite{yuan2013multiobjective} and it incorporated a local search into NSGA-II \cite{deb2002fast}. A hierarchical strategy was adopted in the local search to handle three objectives: makespan, total workload and maximum workload. In section 6, these algorithms have been compared with our algorithm on the MOFJSP.

\subsection{Parameter Tuning}

EA involves using multiple parameters, such as the crossover probability, mutation probability, computational budget, as so on. The preset values of these parameters affect the performance of the algorithm in different situations. The parameters are usually set to values which are assumed to be good. For example, the mutation probability normally is kept very low, otherwise the convergence is supposed to be delayed unnecessarily. But the best way to identify the probability would be to do a sensitivity analysis: carrying out multiple runs of the algorithms with different mutation probabilities and comparing the outcomes. Although there are some self-tuning techniques for adjusting these parameter on the go, the hyper-parameters in EA can be optimized using the technique from machine learning. 

The optimization of hyper-parameters and neural network architectures is a very important topic in the field of machine learning due to the large number of design choices for a network architecture and its parameters. Recently, algorithms have been developed to accomplish this automatically since it is intractable to do it by hand. The MIP-EGO \cite{van2018automatic} is one of these configurators that can automatically configure convolutional neural network architectures and the resulting optimized neural networks have been proven to be competitive with the state-of-the-art manually designed ones on some popular classification tasks. Especially, MIP-EGO allows for multiple candidate points to be selected and evaluated in parallel, which can speed up the automatic tuning procedure. In our paper, we tune several parameters with MIP-EGO to find the best parameter setting for them.

\subsection{NSGA-III}
NSGA-III is a decomposition-based MOEA, it is an extension of the well-know NSGA-II and eliminates the drawbacks of NSGA-II such as the lack of uniform diversity among a set of non-dominated solutions. The basic framework of NSGA-III is similar to the original NSGA-II, while it replaces the crowding distance operator with a clustering operator based on a set of reference points. A widely-distributed set of reference points can efficiently promote the population diversity during the search and NSGA-III defines a set of reference points by Das and Dennis$'$s method \cite{das1998normal}.

In each iteration $t$, an offspring population $Q_t$ of size $N_{pop}$ is created from the parent population $P_t$ of size $N_{pop}$ using usual selection, crossover and mutation. Then a
combined population $R_t$ = $P_t \cup Q_t$ is formed and classified into different layers ($F_1$, $F_2$, and so on ), each layer consists of mutually non-dominated solutions. Thereafter, starting from the first layer, points are put into a new population $S_t$. A whole population is obtained until the first time the size of $S_t$ is equals to or larger than $N_{pop}$. Suppose the last layer included in $S_t$ is the $l$-th layer, so far, members in $S_t\setminus F_l$ are points that have been chosen for $P_{t+1}$ and the next step is to choose the remaining points from $F_l$ to make a complete $P_{t+1}$. In general (when the size of $S_t$ doesn't equal to $N_{pop}$), $N_{pop}-| S_t \setminus F_l|$ solutions from $F_l$ needs to be selected for $P_{t+1}$. 

When selecting individuals from $F_l$, first,
each member in $S_t$ is associated with a reference point by searching the shortest perpendicular distance from the member to all reference lines created by joining the ideal point with reference points. Next, a niching strategy is employed to choose points associated with the least reference points in $P_{t+1}$ from $F_l$. The niche count for each reference point, defined as the number of members in $S_t\setminus F_l$ that are associated with the reference point, is computed. The member in $F_l$ associated with the reference point having the minimum niche count is included in $P_{t+1}$. The niche count of that reference point is then increased by one and the procedure is repeated to fill the remaining population slots of $P_{t+1}$.

NSGA-III is powerful to handle problems with non-linear characteristics as well as having many objectives. Therefore, we decide to enhance NSGA-III in our algorithm for the MOFJSP. 

\section{PROPOSED ALGORITHM}
\label{sec:proposedalgorithm}
The proposed algorithm, \emph{Flexible Job Shop Problem Multi-objective Evolutionary Algorithm} (FJSP-MOEA) can in principal be combined with any MOEA and help MOEAs solve the MOFJSP, whereas the standard MOEAs cannot solve MOFJSP solely. The algorithm follows the flow of a typical EA and generates improved solutions by using local search. Details of the following components are given in the next subsections.
\begin{itemize}
\item  Initialization: encode the individual and generate the initial population.
\item  Genetic operators: generate offspring by crossover and mutation operators.
\item  Local search: decode the individual and improve the solution with local search.
\end{itemize}

\subsection{Initialization}
\subsubsection{Chromosome Encoding}

The MOFJSP is a combination of assigning each operation to a machine and ordering operations on the machines. In the algorithm, each chromosome (individual) represents a solution in the search space and the chromosome consists of two parts: the operation sequence vector and the machine assignment vector. Let $N$ denote the number of all operations of all jobs. The length of both vectors is equal to $N$. The operation sequence vector decides the sequence of operations assigned to each machine. For any two operations which are processed by the same machine, the one located in front is processed earlier than the other one. The machine assignment vector assigns the operations to machines, in other words, it determines which operation is processed by which machine and the machine should be the one capable of processing the operation.

The format of representing an individual not only influences the implementation of crossover and mutation operators, a proper representation can also avoid the production of infeasible schedules and reduces the computational time. In our algorithm, the chromosomal representation proposed by Zhang et al. in \cite{zhang2011effective} is adopted and an example is given in Table~\ref{table_2}.

\begin{table}[htbp]
\caption{An example of a chromosome representation}
\label{table_2}
\begin{center}
\begin{tabular}{|c||c|c|c|c|c|c|c|}
\hline
Operation sequence&  \pmb{1} &  \pmb{2} &  \pmb{3} &  \pmb{2} &  \pmb{1} &  \pmb{1} &  \pmb{3}\\
\hline
 & $O_{11}$ & $O_{21}$ &  $O_{31}$ &  $O_{22}$ & $O_{12}$ &  $O_{13}$ &  $O_{32}$\\
\hline
Machine assignment&  \pmb{2} &  \pmb{1} &  \pmb{1} &  \pmb{3} &  \pmb{2} &  \pmb{2} &  \pmb{1}\\
\hline
 & $O_{11}$ & $O_{12}$ &  $O_{13}$ &  $O_{21}$ & $O_{22}$ &  $O_{31}$ &  $O_{32}$\\
\hline
 & $M_3$ & $M_1$ & $M_3$ & $M_3$ & $M_3$ & $M_2$ & $M_1$\\
\hline
\end{tabular}
\end{center}
\end{table}

In Table~\ref{table_2}, the first row shows the operation sequence vector which consists of only job indexes. For each job, the first appearance of its index represents the first operation of that job and the second appearance of the same index represents the second operation of that job, and so on. The occurrence number of an index is equal to the number of operations of the corresponding job. The second row explains the first row by giving the real operations. The third row is the machine assignment vector which presents the selected machines for all operations. The operation sequence of the machine assignment vector is fixed, which is from the first job to the last job and from the first operation to the last operation for each job. The fourth row indicates the fixed operation sequence of the machine assignment vector and the fifth row shows the real machines of the operations. Each integer value in the machine assignment vector is the index of the machine in the set of alternative machines of that operation. In this example, $O_{13}$ is assigned to $M_3$ because $M_3$ is the first (and only) machine in the alternative machine set of $O_{13}$ (Table~\ref{table_1}). The alternative machine set of $O_{22}$ is $\{ M_1, M_3\}$, the second machine in this set is $M_3$, therefore, $O_{22}$ is assigned to $M_3$. 

\subsubsection{Initial population}

Our algorithm starts by creating the initial population. The machine assignment and operation sequence vectors are generated separately for each individual. In the literature, a few approaches have been proposed for producing individuals, such as global minimal workload in \cite{kacem2002approach}; AssignmentRule1 and AssignmentRule2 in \cite{pezzella2008genetic}. In our algorithm, several new methods are proposed, namely the \emph{Processing Time Roulette Wheel} (PRW) and \emph{Workload Roulette Wheel} (WRW) for initialising the machine assignment and the \emph{Most Remaining Machine Operations} (MRMO) and \emph{Most Remaining Machine Workload} (MRMW) for initialising the operation sequence. These new approaches have used together with some commonly used dispatching rules in initializing individuals on the purpose of enriching the initial population. When generating a new individual in our algorithm, two initialization methods are randomly picked from the following two lists; one for the machine assignment vector and one for the operation sequence vector.

\subsection*{Initialization methods for machine assignment}
\begin{enumerate}[leftmargin=1em,itemindent=0em]

\item Random assignment (Random): an operation is assigned to an eligible machine randomly.

\item Processing time Roulette Wheel (PRW): for each operation, the roulette wheel selection is adopted to select a machine from its machine set based on the processing times of these capable machines. The machine with the shorter processing time is more likely to be selected. 

\item Workload Roulette Wheel (WRW): for each operation, the roulette wheel selection is used to select a machine from its machine set based on the current workloads plus the processing times of these capable machines. The machine with lower sum of the workload and processing time is more likely to be selected.
\end{enumerate}

We propose PRW and WRW to assign the operation to the machine with less processing time or accumulated workload, at the same time, maintaining the freedom of exploring the entire search space. 





\subsection*{Initialization methods for operation sequence}
\begin{enumerate}[leftmargin=1em,itemindent=0em]
\item Random permutation (Random): starting from a fixed sequence: all job indexes of $J_1$ (the number of $J_1$ job indexes is the number of operations of $J_1$), followed by all job indexes of $J_2$, and so on. Then the array with the fixed sequence is permuted and a random order is generated.

\item Most Work Remaining (MWR): operations are placed one by one into the operation sequence vector. Before selecting an operation, the remaining processing times of all jobs are calculated respectively, the first optional operation of the job with the longest remaining processing time is placed into the chromosome.

\item Most number of Operations Remaining (MOR): operations are placed one by one into the operation sequence vector. Before selecting an operation, the number of succeeding operations of all jobs is counted respectively, the first optional operation of the job with the most remaining operations is placed into the chromosome.

\item Long Processing Time (LPT)\cite{xing2009efficient}: operations are placed one by one into the operation sequence vector, each time, the operation with maximal processing time is selected without breaking the order of jobs.

\item Most Remaining Machine Operations (MRMO): operations are placed into the operation sequence vector according to both the number of subsequent operations on machines and the number of subsequent operations of jobs. MRMO is a hierarchical method and takes the machine assignment into consideration. First, the machine with the most subsequent operations is selected. After that, the optional operations in the subsequent operations on that machine are found based on the already placed operations. For example, if $O_{11}\to O_{12}\to O_{21}$ are placed operations, the current optional operation can only be chosen from $O_{13}$, $O_{22}$, and $O_{31}$. In these optional operations, those which are assigned to the selected machine are picked and the one that belongs to the job with the most subsequent operations is placed into the chromosome. In this example, $O_{31}$ will be chosen if it is assigned to the selected machine because there are two subsequent operations for $J_3$ and only one subsequent operation for $J_1$ and $J_2$. Note that it is possible that no operation is available on that machine, in that case, the machine with the second biggest number of subsequent operations will be selected, and so forth.

\item Most Remaining Machine Workload (MRMW): operations are placed into the operation sequence vector according to both the remaining processing times of machines and the remaining processing times of jobs. MRMW is a hierarchical method similar to MRMO. After finding the machine with the longest remaining process time and the optional operations on that machine, the operation which belongs to the job with the longest remaining process time is placed into the chromosome. Again, if no operation is available on that machine, the machine with the second longest remaining processing time will be selected, and so forth.
\end{enumerate}

We propose MRMO and MRMW to give priority to both the machine and the job with the most number of remaining operations (MRMO) and the longest remaining processing time (MRMW). 

\subsection{Crossover}

Crossover is a matter of replacing some of the genes in one parent with the corresponding genes of the other (Glover and Kochenberger \cite{glover1handbook}). Since our representation of chromosomes has two parts, crossover operators applied to these two parts of chromosomes are implemented separately as well. We propose two new crossover operators, \emph{Precedence Preserving Two Points Crossover} (PPTP) and \emph{Uniform Preservative crossover} (UPX), and use them together with several commonly adopted crossover operators. When executing the crossover operation in the proposed algorithm, one crossover operator for machine assignment and one operator for the operation sequence, are randomly chosen from the following two lists to generate the offspring.


\subsection*{Crossover operators for machine assignment}


\begin{enumerate}[leftmargin=1em,itemindent=0em]
\item No crossover 

\item One point crossover: a cutting point is picked randomly and genes after the cutting point are swapped between two parents.

\item Two points crossover: two cutting points are picked randomly and genes between the two points are swapped between two parents.

\item Job-based crossover (JX): it generates two children from two parents by the following procedure:

\begin{itemize}
    \item[a] A vector with the size of the jobs is generated, which consists of random values 0 and 1. 
    \item[b] For the job corresponding to value $0$, the assigned machines of its operations are preserved.
    \item[c] For the job corresponding to value $1$, the machines of its operations are swapped between two parents.
\end{itemize}

\item Multi-point preservative crossover (MPX)\cite{zhang2007bilevel}: MPX generates two children from two parents by the following procedure:

\begin{itemize}
    \item[a] A vector with the size of all operations is generated, which consists of random values $0$ and $1$.
    \item[b] For the operations corresponding to value $0$, their machines (genes) are preserved.
    \item[c] For the operations corresponding to value $1$, their machines (genes) are swapped between the two parents.
\end{itemize}
\end{enumerate}

\subsection*{Crossover operators for operation sequence}

\begin{enumerate}[leftmargin=1em,itemindent=0em]
\item No crossover 

\item Precedence preserving one point crossover (PPOP) \cite{teekeng2012modified}: PPOP generates two children from two parents by the following procedure:

\begin{itemize}
    \item[a] A cutting point is picked randomly, genes to the left are preserved and copied from parent1 to child1 and from parent2 to child2.
    \item[b] The remaining operations in parent1 are reallocated in the order they appear in parent2.
    \item[c] The remaining operations in parent2 are reallocated in the order they appear in parent1.
\end{itemize}

An example of PPOP is shown in  Figure~\ref{figure_1} and  the cutting point is between the third and fourth operation. Red numbers in parent2 are the genes on the right side of the cutting point in parent1 and they are copied to child1 with their own sequence following the genes on the left side of the cutting point in parent1, and vice versa.

\begin{figure}[thpb]
\hspace{-0.5cm}
   \centering
         \includegraphics[scale=0.4]{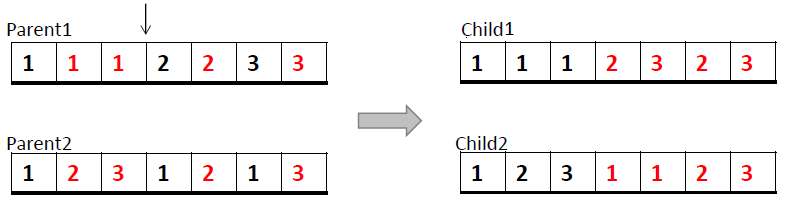}
    \caption{The process of PPOP}
    \label{figure_1}
\end{figure}

\item Precedence Preserving Two Points Crossover (PPTP): PPTP generates two children from two parents by the following procedure:

\begin{itemize}
    \item[a] Two cutting points are picked randomly, genes except for those between the two points are preserved and copied from parent1 to child1 and from parent2 to child2.
    \item[b] Operations between the two cutting points in parent1 are reallocated in the order they appear in parent2.
    \item[c] Operations between the two cutting points in parent2 are reallocated in the order they appear in parent1.
\end{itemize}

\item Improved precedence operation crossover (IPOX)\cite{zhang2005new}: IPOX divides the job set into two complementary and non-empty subsets randomly. The operations of one job subset are preserved, while the operations of another job subset are copied from another parent.

\item Uniform Preservative crossover (UPX): UPX generates two children from two parents by the following procedure:

\begin{itemize}
    \item[a] A vector with the size of all operations is generated, which consists of random values $0$ and $1$.
    \item[b] For the operations corresponding to value $0$, the genes are preserved and copied from parent1 to child1 and from parent2 to child2.
    \item[c] For the operations corresponding to value $1$, the genes in parent1 are found in parent2 and copied from parent2 with the sequence in parent2, and vice versa.
\end{itemize}

\end{enumerate}

\subsection{Mutation}

The mutation operator flips the gene values at selected locations. By forcing the algorithm to search areas other than the current area, the mutation operator is used to maintain genetic diversity from one generation of a population to the next. In our algorithm, insertion mutation and swap mutation (including one point swap and two points swap) are proposed and used.

\textbf{Insertion Mutation Operator} generates a new individual by the following procedure:
    \begin{itemize}
        \item Two random numbers $i$ and $j$ ($1\leq i \leq N$,  $1\leq j \leq N$) are selected.
        \item For the operation sequence vector, the operation on position $j$ is inserted in front of the operation on position $i$. 
        \item For the machine assignment vector, a machine is randomly selected for both the operations on $i$ and on $j$ respectively. If the processing time on the newly selected machine is lower than that on the current machine, the current machine is replaced by the new machine. If the processing time on the new machine is longer than that on the old machine, there is only a 20\% probability that the new machine replaces the old machine.
    \end{itemize}
    
\textbf{Swap Mutation Operator} generates a new individual by the following procedure:
    
    \begin{itemize}
        \item One random number $i$ ($1\leq i \leq N$) is selected or two random numbers $i$ and $j$ ($1\leq i \leq N$,  $1\leq j \leq N$) are selected.
        \item For the operation sequence vector, with only one swap point $i$, the operation on the swap point is swapped with its neighbour; with two swap points, the operations on position $i$ and $j$ are swapped.
        \item For the machine assignment vector, the machine on position $i$ (and $j$) is replaced with a new machine by the same rule used in the insertion mutation operator.
    \end{itemize}

\subsection{Decoding and Local Search}

Decoding a chromosome is to convert an individual into a feasible schedule to calculate the objective values which represents the relative superiority of a solution. In this process, the operations are picked one by one from the operation sequence vector and placed on the machines from the machine assignment vector to form the schedule. When placing each operation to its machine, local search (in the sense of heuristic rules to improve solution) is involved to refine an individual in order to obtain an improved schedule in the proposed algorithm. Two levels of local search are applied to allocate each operation to a time slot on its machine. We know that idle times may exist between operations on each machine due to precedence constraints among operations of each job, and two levels of local search utilize idle times in different degrees.

\subsubsection*{The first level local search}
\begin{figure}[htbp]
\begin{minipage}[t]{0.5\textwidth}
\hspace{-0.2cm}
\includegraphics[width=2.5in]{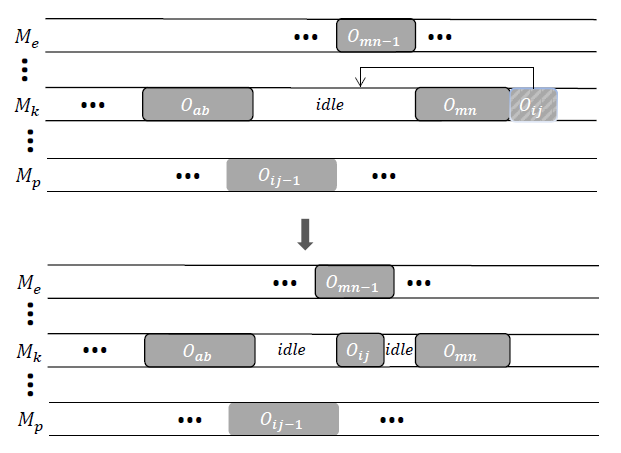}
\caption{First level local search}\label{figure_2}
\end{minipage}
\begin{minipage}[t]{0.5\textwidth}
\hspace{0cm}
\includegraphics[width=2.4in]{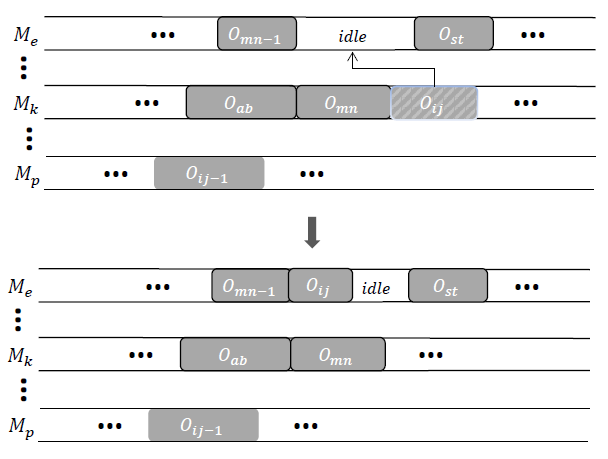}
\caption{Second level local search}\label{figure_3}
\end{minipage}
\end{figure}

let $S_{ij}$ be the starting time of $O_{ij}$ and $C_{ij}$ the completion time of $O_{ij}$, an example of the first level local search is shown in Figure~\ref{figure_2}. Because $O_{mn}$ needs to be processed after the completion of $O_{mn-1}$, an idle time interval between the completion of $O_{ab}$ and the starting of $O_{mn}$ appeared on machine $M_k$. $O_{ij}$ is assigned to $M_k$ and we assume that $O_{mn}$ is the last operation on $M_k$ before handling $O_{ij}$, therefore the starting time of $O_{ij}$ is $\max \{ C_{mn}, C_{ij-1}  \}$, which in this example is $C_{mn}$ and it is later than $C_{ij-1}$, thus, there is an opportunity that $O_{ij}$ can be processed earlier. When checking the idle time on $M_k$, the idle time interval $[C_{ab}, S_{mn}]$ is found available for $O_{ij}$ because the idle time span $[C_{ij-1}, S_{mn}]$, which is part of $[C_{ab}, S_{mn}]$, is enough to process $O_{ij}$ or longer than $t_{ijk}$.

Let $S^d_{k}$ be the starting time of the $d$th idle time interval on $M_k$ and $C^d_{k}$ be the completion time. $O_{ij}$ can be transferred to an earliest possible idle time interval of its machine which satisfies the following equation:

\begin{equation}\label{eq:4}
    \max\{S^d_{k}, C_{ij-1}\}+t_{ijk} \leq C^d_{k},  (C_{ij}=0,  \mbox{if}~ j=1)
\end{equation}

After using the idle time interval, the starting time of $O_{ij}$ is $\max\{S^d_{k}, C_{ij-1}\}$ and the idle interval is updated based on the starting and completion time of $O_{ij}$: (1) the idle time interval is removed; (2) the starting or completion time of the idle time interval is modified; (3) the idle time interval is replaced by two new shorter idle time intervals, like in the example of Figure~\ref{figure_2}.

After decoding a chromosome, the operation sequence vector of the chromosome is updated according to new starting times of operations, and three objective values are calculated. The first level local search only finds for each operation the available idle time interval on its assigned machine. After generating the corresponding schedule with the first level search method, it is possible that there are still operations that can be allocated to available idle time intervals to benefit the fitness value. To achieve this, decoding the chromosome which has been updated with the first level local search is performed with the second level local search, and again operations are moved to available idle time intervals. 

\subsubsection*{The second level local search}

The second level local search not only checks the idle time intervals on the assigned machine, but also the idle time intervals on alternative machines. An example of making use of the idle time interval on another machine is shown in Figure~\ref{figure_3}. Let $S_{ijk}$ be the starting time and $C_{ijk}$ be the completion time of $O_{ij}$ on $M_k$. In this example, $O_{ij}$ is assigned to $M_k$ in the initial chromosome, we assume that $O_{ij}$ can also be performed by  $M_e$. Under the condition that the starting time of $O_{ij}$ on $M_k$ is later than the completion time of $O_{ij-1}$, the idle time intervals on all alternative machines which can process $O_{ij}$ are checked. An idle time interval on $M_e$ could be a choice and $O_{ij}$ can be reallocated to $M_e$. In this example, the processing time of $O_{ij}$ on $M_e$ is even shorter then the processing time on $M_k$, therefore, this reallocation can at least benefit the total workload.

In the second level local search, all available idle time intervals of an operation are checked one by one until the first ``really" available idle time interval is found and then the operation is moved to that idle time interval. Any idle time interval on an alternative machine which can satisfy Equation \ref{eq:4} is an available idle time interval, while it must meet at least one of the following conditions to become a ``really" available idle time interval.
\begin{itemize}
\item[1.] The processing time of the operation on the new machine is shorter than on the initially assigned machine if the available idle time interval is on a different machine;

\item[2.] The operation can be moved from the machine with the maximal makespan to another machine.

\item[3.] The operation can be moved from the machine with the maximal workload to another machine.
\end{itemize}
The total workload can be improved directly by the first condition; the motive of the second condition is to decrease the maximal makespan and the third condition can benefit the critical workload.

After the reallocation of the operations with the second level local search, the corresponding schedule is obtained and objective values are calculated. While, instead of updating the chromosome immediately, the new objective values are compared with the old objective values first, the chromosome can be updated only when at lease one objectives is better than its old value. This is to make sure that the new schedule is at least not worse than the old schedule (The new solution is not dominated by the old solution).
Another difference between the first and second level local search is that the first level local search is performed on every evaluation, while the second level local search is only performed with a 30\% probability for each chromosome to avoid local optima. Although these two local searches can be applied repeatedly to improve the solution, to avoid that the algorithm is stuck in a local optima, they are employed only once for each evaluation.



\section{Experiments and results}
\label{sec:experiments}

The experiments are implemented on the MOEA Framework (version 2.12, available from http://www.moeaframework.org). The algorithms are tested on two sets of well-known FJSP benchmark instances: 4 Kacem instances (ka4x5, ka10x7, ka10x10, ka15x10) and 10 BRdata instances (Mk01-Mk10). Table~\ref{table_4} gives the scale of these instances. The first column is the name of each instance; the second column shows the size of the instance, in which $n$ stands for the number of jobs and $m$ the number of machines; the third column represents the number of operations; the fourth column lists the flexibility of each instance, which means the average number of alternative machines for each operation in the problem.

\begin{table}[h]
\caption{The scale of benchmark instances}
\label{table_4}
\begin{center}
\begin{tabular}{|c|c|c|c|}
\hline
Instance & n $\to$ m & \#Opr & Flex.\\
\hline
ka4x5 & 4 $\to$ 5 & 12 & 5 \\
\hline
ka10x7 & 10 $\to$ 7 & 29 & 7 \\
\hline
ka10x10 & 10 $\to$ 10 & 30 & 10\\
\hline
ka15x10 & 15 $\to$ 10 & 56 & 10\\
\hline
Mk01 & 10 $\to$ 6 & 55 & 2 \\
\hline
Mk02 & 10 $\to$ 6 & 58 & 3.5 \\
\hline
Mk03 & 15 $\to$ 8 & 150 & 3 \\
\hline
Mk04 & 15 $\to$ 8 & 90 & 2 \\
\hline
Mk05 & 15 $\to$ 4 & 106 & 1.5 \\
\hline
Mk06 & 10 $\to$ 15 & 150 & 3 \\
\hline
Mk07 & 20 $\to$ 5 & 100 & 3 \\
\hline
Mk08 & 20 $\to$ 10 & 225 & 1.5\\
\hline
Mk09 & 20 $\to$ 10 & 240 & 3 \\
\hline
Mk10 & 20 $\to$ 15 & 240 & 3 \\
\hline
\end{tabular}
\end{center}
\end{table}

All the experiments are performed with a population size of $100$, each run of the algorithm will stop based on a predefined number of evaluation, which is $10,000$ for Kacem instances and $150,000$ for BRdata instances. For each problem instance, the proposed algorithm is independently run $30$ times. The resulting solution set of an instance is formed by merging all the non-dominated solutions from its $30$ runs. 

The crossover probability is set to $1$ and two random crossover operators can be chosen each time (one for operation sequence and one for machine assignment). For Kacem instances, the mutation probabilities are set to $0.6$. For BRdata instances, which include larger-scale and more complex problems, the MIP-EGO configurator \cite{van2018automatic} is adopted to tune both insertion and swap mutation probabilities (one point swap mutation and two points swap mutation) to find the best parameter values for each problem. The hypervolume of the solution set has been used in MIP-EGO as the objective value to tune three mutation probabilities. Although the true Pareto fronts (PF) for test instances are unknown, \cite{yuan2013multiobjective} provides the reference set for Kacem and BRdata FJSP instances, which is formed by gathering all non-dominated solutions found by all the implemented algorithms in \cite{yuan2013multiobjective} and also non-dominated solutions from other state-of-the-art MOFJSP algorithms. 
We define the reference point for calculating the hypervolume value based on the largest value in this reference set. To be specific, each objective function value of the reference point is: $1.1 ~\times$ largest objective function value of the respective dimension in the reference set. The origin point is used as the ideal point. Other basic parameter settings of MIP-EGO are listed in Table~\ref{table_5}. For each mutation probability, we only consider a discretized number with only one digit after the decimal point, therefore, the search space is ordinal or integer space, which in MIP-EGO are handled in the same way.
\begin{table}[htbp]
\caption{Settings for MIP-EGO}
\label{table_5}
\begin{center}
\begin{tabular}{l|c}
\hline
Parameter & value\\
\hline
maximal number of evaluations & 200\\
surrogate model &  random forest\\
optimizer for infill criterion & MIES\\
search space & ordinal space\\
\hline
\end{tabular}
\end{center}
\end{table}


With a budget of $200$ evaluations, Table~\ref{table_x} shows the percentage of the evaluations which can achieve the largest hypervolume value (or the best PF) by MIP-EGO. It can be observed  for Mk05 and Mk08 that all the evaluations have obtained the largest hypervolume value, it means that all parameter values of mutation probabilities in MIP-EGO can achieve the best PF for these two problems. It can also be seen in Table~\ref{table_4} that both problems have a low flexibility value. On the contrary, for Mk06, Mk09 and Mk10, these problems have a large operation number and high flexibility.  It seems that they can be difficult to solve because there is only one best parameter setting for the mutation probabilities. This also means that it is highly likely better solution sets can be found with a higher budget.

\begin{table}[htbp]
\caption{Probability of finding best configuration}
\label{table_x}
\begin{center}
\begin{tabular}{|c|c|c|c|c|c|c|c|c|c|}
\hline
Mk01 & Mk02 & Mk03 & Mk04 & Mk05 & Mk06 & Mk07 & Mk08 & Mk09 & Mk10\\
\hline
 $73\%$ & $60\%$ &  $95\%$ &  $1\%$ & $100\%$ & $0.5\%$ & $4.5\%$ &  $100\%$ &  $0.5\%$ & $0.5\%$\\
\hline
\end{tabular}
\end{center}
\end{table}

With the best parameter setting of the mutation probabilities for BRdata instances, we compared our experimental results with the reference set in \cite{yuan2013multiobjective}. Our algorithm can achieve the same Pareto optimal solutions as in the reference set for all BRdata instances except for Mk06, Mk09 and Mk10. At the same time, for Mk06 and Mk10, our algorithm can find new non-dominated solutions. Table~\ref{table_newsolutions} is the list of new non-dominated solutions obtained by our algorithm, each row of an instance is a solution with three objectives: makespan, total workload, and critical workload. 

\begin{table}[htbp]
\caption{Newly achieved non-dominated solutions}
\vspace{-0.1cm}
\label{table_newsolutions}
\begin{center}
\begin{tabular}{|c c c|c c c|}
\hline
\multicolumn{3}{|c|}{Mk06}& \multicolumn{3}{c|}{Mk10}\\
\hline
61 &  427 &  53 & 218 & 1973 & 195\\
63 &  428 &  52 & 218 & 1991 & 194\\
63 &  435 &  51 & 219 & 1965 & 195\\
65 &  453 &  49 & 220 & 1984 & 191\\
66 &  451 &  49 & 225 & 1979 & 194\\
66 &  457 &  48 & 226 & 1954 & 196\\
 &   &   & 226 & 1974 & 194\\
 &   &   & 226 & 1979 & 192\\
 &   &   & 228 & 1973 & 194\\
 &   &   & 235 & 1938 & 199\\
 &   &   & 236 & 1978 & 193\\
\hline
\end{tabular}
\end{center}
\end{table}

\vspace{-0.1cm}
Another comparison is between our algorithm (FJSP-MOEA) and MOGA \cite{wang2010multi}, SEA \cite{chiang2013simple} and MA1, MA2 \cite{yuan2013multiobjective}. In \cite{yuan2013multiobjective}, there are several variants of the proposed algorithm with different strategies in the local search. We pick MA1 and MA2 as compared algorithms because they perform equally good or superior to other algorithms on almost all problems.  
Table~\ref{table_hv} displays the hypervolume value of the PF approximation from all algorithms and the new reference set which is formed by combining all solutions from the PF by all algorithms. The highest hypervolume value on each problem in all algorithms has been highlighted in bold. We observed that FJSP-MOEA and MA1, MA2 show the best and similar performance, and MOGA behaves the best for three of the BRdata instances. The good performance of MOGA on three problems is interesting. MOGA has a entropy-based mechanism to maintain decision space diversity which might be beneficial for solving these problem instances. When using one best parameter setting, we also give the average hypervolume and standard deviation from 30 runs on each problem in Table~\ref{ave-hv}, the standard deviation of each problem shows the stable behaviour of each run.  

\begin{table}[h]
\caption{Hypervolume from MOGA, SEA, MA1, MA2, FJSP-MOEA and the reference set}
\label{table_hv}
\begin{center}
\begin{tabular}{|c|c|c|c|c|c|c|}
\hline
Problem & MOGA & SEA & MA1 & MA2 & FJSP-MOEA & Ref\\
\hline
Mk01 & 0.00426 & 0.00508 & \bfseries{0.00512}& \bfseries{0.00512} & \bfseries{0.00512} &0.00512\\
\hline
Mk02 & 0.01261 & 0.01206 & \bfseries{0.01294} &\bfseries{0.01294} &\bfseries{0.01294} &0.01294\\
\hline
Mk03 & \bfseries{0.02460} & 0.02165 & 0.02165 & 0.02165 & 0.02165 & 0.02809\\
\hline
Mk04 & \bfseries{0.06906} & 0.06820 & 0.06901 & 0.06901 & 0.06901 & 0.07274 \\
\hline
Mk05 & 0.00626 & 0.00635 & \bfseries{0.00655} & \bfseries{0.00655} & \bfseries{0.00655} & 0.00655\\
\hline
Mk06 & 0.05841 & 0.06173 & 0.06585 & 0.06692 & \bfseries{0.06709} & 0.07065\\
\hline
Mk07 & 0.02244 & 0.02132 &\bfseries{0.02269} &\bfseries{0.02269} & \bfseries{0.02269} & 0.02288\\
\hline
Mk08 & \bfseries{0.00418} & 0.00356 & 0.00361 & 0.00361 & 0.00361 & 0.00428\\
\hline
Mk09 & 0.01547 & 0.01755 &  0.01788 &  \bfseries{0.01789} & 0.01785 & 0.01789 \\
\hline
Mk10 & 0.01637 & 0.01778 & 0.02145 & \bfseries{0.02196} & 0.02081 & 0.02249\\
\hline
\end{tabular}
\end{center}
\end{table}

\begin{table}[htbp]
\caption{Average hypervolume and std with the best parameter setting}
\label{ave-hv}
\begin{center}
\begin{tabular}{|c|c|c|c|c|c|c|c|c|c|c|}
\hline
Problem &Mk01 & Mk02 & Mk03 & Mk04 & Mk05 & Mk06 & Mk07 & Mk08 & Mk09 & Mk10\\
\hline
Average HV& $0.0050$ & $0.0122$ &  $0.0216$ &  $0.0672$ & $0.0064$ & $0.0598$ & $0.0222$ &  $0.0036$ &  $0.0174$ & $0.0186$\\
\hline
Std& $0$ & $0.0003$ &  $0.0001$ &  $0.0004$ & $0.0001$ & $0.0019$ & $0.0003$ &  $0$ &  $0.0002$ & $0.0006$\\
\hline
\end{tabular}
\end{center}
\end{table}

For Kacem instances and with fixed mutation probabilities, our obtained non-dominated solutions are the same as the PF in the reference set. MA1 and MA2 also achieved the best PF for all Kacem instances, but our algorithm uses far less computational resources. The proposed FJSP-MOEA uses only a population size of $100$ whereas the population size of MA algorithms is $300$. FJSP-MOEA uses only $10,000$ objective function evaluations, whereas MA used $150,000$ evaluations. In terms of computational resources the proposed FJSP-MOEA can therefore be used on smaller computer systems, entailing broader applicability, and possibly also in real-time algorithm implementations such as dynamic optimization.

\section{CONCLUSIONS}
\label{sec:conclusions}
A novel multi-objective evolutionary algorithm for MOFJSP is proposed. It uses multiple initialization approaches to enrich the first generation population, and various crossover operators to create better diversity for offspring. Moreover, to determine the optimal mutation probabilities, the MIP-EGO configurator is adopted to automatically generate proper mutation probabilities. Besides, the straightforward local search is employed with different levels to aid more accurate convergence to the PF. The proposed customization approach in principle can be combined with almost all MOEAs. In this paper, we incorporate it with one of the state-of-the-art MOEAS, namely NSGA-III, to solve MOFJSP, and the new algorithm can find all Pareto optimal solutions in literature for most problems, and even new Pareto optimal solutions for the large scale instances.

In this paper, we show the ability of MIP-EGO in finding the optimal mutation probabilities. However, there is more potential in the automated parameter configuration domain that can benefit EA. For example, to know the effects of different initialization approaches and crossover operators, we can optimize the initialization and crossover configuration. Furthermore, other parameters of the proposed algorithm, such as, population size, evaluation number, and so on, can also be tuned automatically. However, so far the efficiency of the existing tuning framework is limited when it comes to a larger number of parameters. It would therefore be a good topic of future research to find more efficient implementations of these. Finally, based on the good performance of MOGA on some of the problems, it seems to be interesting for future research to integrate the entropy-based selection mechanism also into the MOEA schemes to achieve an even better performance.

%
%
%
%

\end{document}